\documentclass[10pt,twocolumn,letterpaper]{article}

\usepackage{cvpr}              

\usepackage{graphicx}
\usepackage{color}
\usepackage{xcolor}
\usepackage{multirow}
\usepackage{booktabs}
\usepackage[super]{nth}
\usepackage{array}
\usepackage{relsize}
\usepackage{wrapfig}
\usepackage{algorithm,algpseudocode}
\usepackage{amsmath}
\usepackage{amsthm}
\usepackage{amssymb}
\usepackage{paralist}
\usepackage{enumitem}
\usepackage{wrapfig}

\usepackage[pagebackref,breaklinks,colorlinks]{hyperref}

\usepackage[capitalize]{cleveref}
\crefname{section}{Sec.}{Secs.}
\Crefname{section}{Section}{Sections}
\Crefname{table}{Table}{Tables}
\crefname{table}{Tab.}{Tabs.}

\newtheorem{remark}{Remark}

\newtheorem{task}{Task}

\crefname{definition}{Definition}{Definitions}
\crefname{assumption}{Assumption}{Assumptions}
\crefname{theorem}{Theorem}{Theorems}
\crefname{remark}{Remark}{Remarks}
\crefname{lemma}{Lemma}{Lemmas}
\crefname{corollary}{Corollary}{Corollaries}
\crefname{proposition}{Proposition}{Propositions}
\crefname{section}{Section}{Sections}
\crefname{subsection}{Subsection}{Subsections}
\crefname{example}{Example}{Examples}
\crefname{table}{Table}{Tables}
\crefname{problem}{Problem}{Problems}
\crefname{algorithm}{Algorithm}{Algorithms}
\crefname{figure}{Figure}{Figures}
\crefname{property}{Property}{Properties}

\def\cvprPaperID{1771} 
\def\confName{CVPR}
\def\confYear{2023}

\begin{document}

\title{Deep Graph Reprogramming}

\author{
Yongcheng Jing$^1$,
Chongbin Yuan$^2$,
Li Ju$^2$,
Yiding Yang$^2$,
Xinchao Wang$^2$,
Dacheng Tao$^1$
 \\ 
$^1$The University of Sydney, Australia,
$^2$National University of Singapore, Singapore\\
{\tt\small
xinchao@nus.edu.sg,
dacheng.tao@gmail.com
}
}

\maketitle

\begin{abstract}
In this paper, we explore a novel model reusing task tailored for graph neural networks (GNNs), termed as ``deep graph reprogramming''.
We strive to reprogram a pre-trained GNN, without amending raw node features nor model parameters, to handle a bunch of cross-level downstream tasks in various domains.
To this end, we propose an innovative Data Reprogramming paradigm alongside a Model Reprogramming paradigm.
The former one aims to address the challenge of diversified graph feature dimensions
for various tasks on the input side, while the latter alleviates the dilemma of fixed per-task-per-model behavior on the model side.
For data reprogramming, we specifically devise an elaborated Meta-FeatPadding method to deal with heterogeneous input dimensions, and also develop a transductive Edge-Slimming as well as an inductive Meta-GraPadding approach for diverse homogenous samples.
Meanwhile, for model reprogramming, we propose a novel task-adaptive Reprogrammable-Aggregator, to endow the frozen model with larger expressive capacities in handling cross-domain tasks.
Experiments on fourteen datasets
across node/graph classification/regression, 3D object recognition, and distributed action recognition,
demonstrate that the proposed methods
yield gratifying results, on par with those by re-training from scratch. 
\end{abstract}

\section{Introduction}
\label{sec:intro}

With the explosive growth of graph data, graph neural networks (GNNs) have been deployed across increasingly wider areas \cite{yang2021spagan,yang2020factorizable,jing2021meta,jing2022learning,yang2021learning}, such as recommendation system \cite{wu2020graph} 
and autonomous driving \cite{sarlin2020superglue,wang2021object,wen2021learning}.
However, the favorable performance for such applications generally comes at the expense of tremendous training efforts and high memory loads, precluding the deployment of GNNs on the edge side.
As such, reusing pre-trained GNNs to alleviate training costs has recently emerged as a trending research topic \cite{feng2022freekd,yang2020distillating,yang2022factorizing,wang2021online,yang2022deep,deng2021graph,jing2021amalgamating,zhou2021distilling,joshi2021representation}.

\begin{figure}[t]
  \centering
  \includegraphics[width=0.475\textwidth]{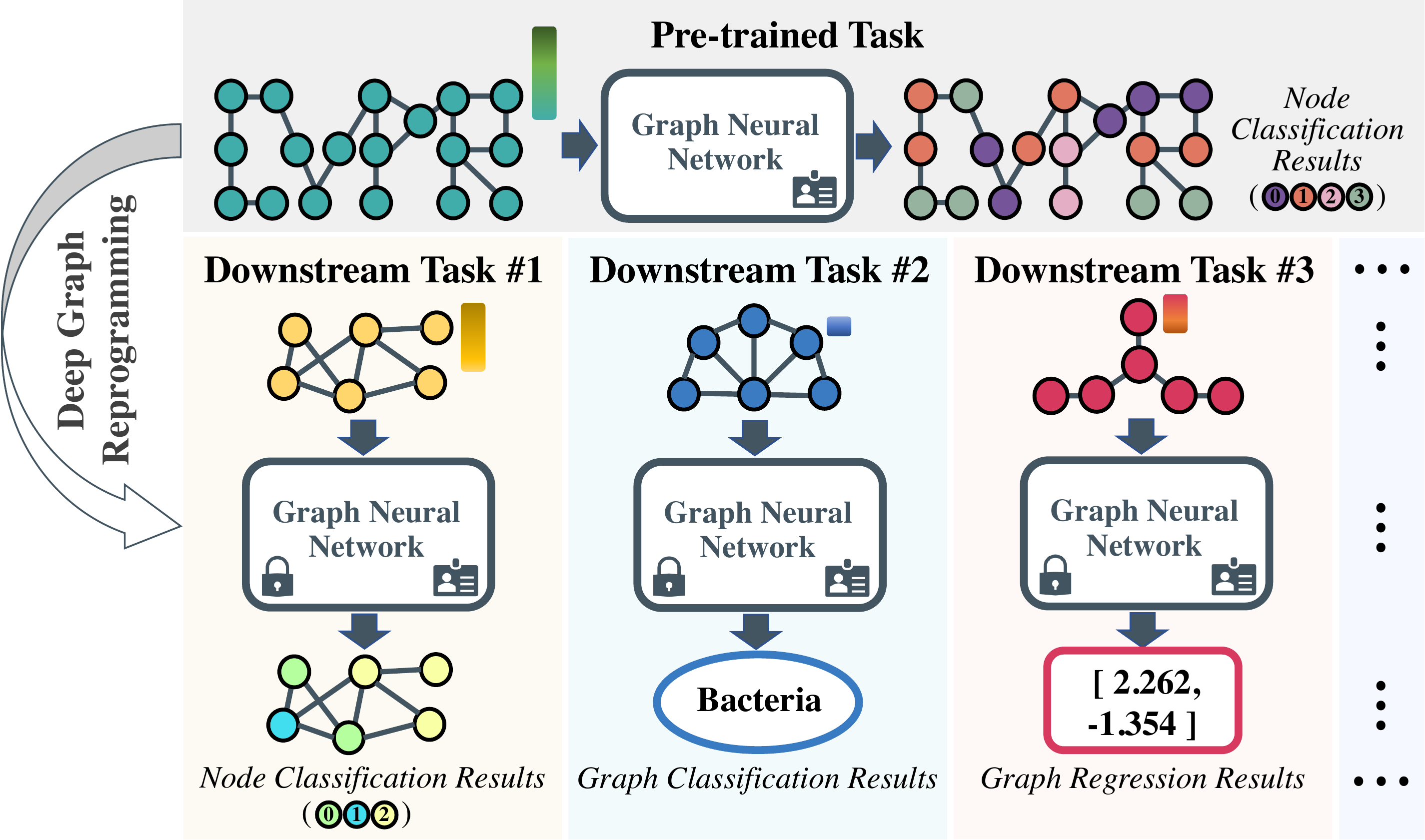}
  \vspace{-5mm}
  \caption{Illustrations of the proposed task of \emph{deep graph reprogramming} (\textsc{Gare}) that aims to reuse pre-trained GNNs to handle plenty of cross-level tasks with heterogeneous graph feature dimensions, without changing model architectures nor parameters.}
  \label{fig:intro}
\end{figure}

Pioneered by the work of \cite{yang2020distillating} that generalize knowledge distillation \cite{hinton2015distilling,Ye2020DataFreeKA,romero2014fitnets,Ye2022LearningWR,song2022spot,Ye2021SafeDB} to the non-Euclidean domain,
almost all existing approaches on reusing GNNs are achieved by following the distillation pipeline in \cite{yang2020distillating}.
Despite the encouraging results, the distilling-based scheme is limited to the \emph{per-task-per-distillation} setting, where a distilled model can only tackle the same task as the teacher can, leading to considerable storage and computation burdens,
especially for the deployment of multiple tasks.

Meanwhile, the distillation mechanism rests upon the hypothesis that abundant pre-trained models are available in the target domains, which indeed holds for image-based areas that always take data in the regular RGB form, thereby readily allowing for \emph{per-model-multiple-dataset} reusing.
However, such an assumption is typically \emph{not} satisfied in the non-Euclidean domain:
on the input side, irregular graph samples have heterogeneous feature dimensions, as shown with the color bars in Fig.~\ref{fig:intro}; on the task side, graph analysis takes various task levels and 
settings, such as graph-, node-, and edge-level learning, as well as transductive and inductive scenarios. 
Such nature of topological diversities leads to inadequate pre-trained GNNs that fit the target downstream tasks.

In this paper, we strive to take one step towards generalized and resource-efficient GNN reusing,
by studying a novel \emph{deep graph reprogramming} (\textsc{Gare}) task. 
Our goal is to reuse a \emph{single} pre-trained GNN across \emph{multiple} task levels and domains, for example the pre-trained one working on node classification and the downstream ones on graph classification and regression, as shown in Fig.~\ref{fig:intro}.
We further impose two constraints to both data and model, where raw features and parameters are frozen in handling downstream tasks.
As such, unlike distillation that essentially
leverages a pre-trained teacher to guide the \emph{re-training} of a student,
the proposed task of \textsc{Gare}, without re-training nor fine-tuning, can thereby be considered to \emph{reprogram} a pre-trained GNN to perform formerly unseen tasks. 

Nonetheless, such an ambitious goal is accomplished with challenges: diversified graph feature dimensions and limited model capacities with a single frozen GNN.
Driven by this observation, we accordingly reformulate \textsc{Gare} into two dedicated paradigms on data and model sides, respectively, termed as \emph{data reprogramming} (\textsc{Dare}) and \emph{model reprogramming} (\textsc{Mere}).
The goal of \textsc{Dare} is to handle downstream graph samples with both the heterogeneous and homogenous dimensions, without amending pre-trained architectures. 
Meanwhile, \textsc{Mere} aims to strengthen the expressive power of frozen GNNs by dynamically changing model behaviors depending on various tasks.  

Towards this end, we propose a universal \emph{Meta-FeatPadding (MetaFP)} approach for heterogeneous-\textsc{Dare} that allows the pre-trained GNN to manipulate heterogeneous-dimension graphs, by accommodating pre-trained feature dimensions via adaptive feature padding in a task-aware manner.
The rationale behind the proposed \emph{MetaFP}, paradoxically, is derived from  \emph{adversarial reprogramming examples} \cite{elsayed2018adversarial} that are conventionally treated as 
attacks to learning systems, where attackers secretly repurpose the use of a target model without informing model providers, by inserting perturbations to input images.
Here we turn the role of the adversarial reprogramming example on its head, by padding around graph perturbations for generalized cross-task model reusing.

Complementary to the dedicated \emph{MetaFP} that is tailored for heterogeneous-\textsc{Dare}, we also devise a \emph{transductive Edge-Slimming (EdgSlim)} and an \emph{inductive Meta-GraPadding (MetaGP)} methods for homogenous-\textsc{Dare}, that handle the downstream graphs with homogenous dimensions under transductive and inductive task settings, respectively, by adaptively eliminating node connections or inserting a tiny task-specific graph, with only, for example, ten vertices, to the raw input sample.
Furthermore, we perform a pilot study on \textsc{Mere}, exploring the pre-trained model capacity for various downstream tasks, by only reprogramming the pre-trained aggregation
behavior (\emph{ReAgg}) upon the well-established \emph{Gumbel-Max} trick.

In sum, our contribution is a novel GNN-based model reusing paradigm that allows for the adaption of a pre-trained GNN to multiple cross-level downstream tasks, and meanwhile requires no re-training nor fine-tuning. 
This is typically achieved through a series of complementary approaches entitled \emph{MetaFP}, \emph{EdgSlim}, and \emph{MetaGP}, that tackle the heterogeneous- and homogenous-dimension graphs within the transductive and inductive scenarios, respectively, together with an elaborated \emph{ReAgg} method to enhance the model capacity.
Experimental results on fourteen benchmarks demonstrate that a pre-trained GNN with \textsc{Gare} is competent to handle all sorts of downstream tasks.

\section{Related Work}
\label{sect:relatedwork}

\noindent
\textbf{Model Reusing.}
With the increasing number of pre-trained GNNs that have been generously released online for reproducibility,
the vision community \cite{chen2021pre,zhaiexploring,wang2021fp,zhai2022one,yang2017pairwise,zhai2020deep} has witnessed a growing interest in reusing GNNs to enhance performance, alleviate training efforts, and improve inference speed \cite{feng2022freekd,joshi2021representation,liu2022dataset,liu2022dynast,fang2023depgraph,wang2022towards,yu2023dataset,wang2023learning}.
The seminal reusing work is performed by Yang \etal \cite{yang2020distillating}, where a dedicated knowledge distillation (KD) method, tailored for GNNs, is proposed to obtain a lightweight GNN from a teacher. 
The follow-up work \cite{deng2021graph} further polishes \cite{yang2020distillating} with a more challenging setting of graph-free KD.
Unlike prior works that merely study KD-based GNN reusing, we propose in this paper a novel parallel task of \textsc{Gare}-based model reusing.

\noindent
\textbf{Universal Model.}
Other than model reusing, graph reprogramming is also essentially a problem of deriving a \emph{universal} model that is applicable to various-domain tasks.
Such universal models have been previously studied in the image and language domains \cite{bommasani2021opportunities,silver2021reward,reed2022generalist,yu2019universally,mccann2018natural}.
In this work, we perform a pilot study on developing universal models in the non-Euclidean domain, 
thereby making one step further towards artificial general intelligence (AGI) \cite{mclean2021risks,jing2023segment}.

\noindent
\textbf{Adversarial Reprogramming.}
The name of the proposed deep graph reprogramming stems from \emph{adversarial reprogramming}, which is a novel type of adversarial attack that seeks to utilize a class-agnostic perturbation to repurpose a target model to perform the task designated by the attacker.
Adversarial reprogramming has recently been studied in various areas, including image classification \cite{elsayed2018adversarial,englert2022adversarial,zheng2021adversarial,kloberdanz2021improved,chen2022model}
and language understanding \cite{hambardzumyan2021warp}.
However, the existence of adversarial reprogramming has not yet been validated in the non-Euclidean graph domain.
Our paper is the first work that explores adversarial reprogramming in GNNs, and further innovatively \emph{turns} such adversarial manner into guards to derive a novel task for resource-efficient model reusing.

\section{Motivation and Pre-analysis}
In this section, we start by giving a detailed analysis on the dilemma of the prevalent reusing scheme of knowledge distillation, and accordingly propose the novel task of \emph{deep graph reprogramming (\textsc{Gare})}, which leads to resource-efficient and generalized GNN reusing.
Then, we uncover the two key challenges of \textsc{Gare} and introduce the proposed paradigms of \textsc{Dare} and \textsc{Mere} with the elaborated rationales on how to address the two challenges.

\subsection{Task Motivation and Definition}

\noindent\textbf{Prevalent Distillation-based Reusing.}
In the literature, almost all existing methods for reusing GNNs are achieved by knowledge distillation (\emph{KD}) 
elaborated in Task~\ref{task:1}.
\vspace{-0.6em}
\begin{task}[\textbf{Reusing GNNs via Knowledge Distillation}]
  \label{task:1}
  \textit{The goal of knowledge distillation is to re-train a compact student model from scratch, that masters the expertise of the pre-trained teacher, via extracting and transferring the knowledge from the pre-trained cumbersome teacher model.}
\end{task}
\vspace{-0.6em}

Such a \emph{KD} manner is inevitably limited by two issues:\\
  \textbf{\large{-}} \textbf{1.} \textbf{\emph{KD}} is built upon an ideal condition that for any downstream task, sufficient pre-trained teacher models are always available for reusing. Such an assumption indeed holds for most cases of image analysis, where the input data is always RGB-pattern. As such, the publicly available model trained on large-scale datasets, such as \emph{ImageNet}, is readily reusable for downstream classification tasks. However, graph data instead has highly diversified input dimensions, feature types (\eg, node and edge features), as well as various task levels (\eg, node- and graph-level analysis), making it challenging to reuse online-released GNNs, as image-domain does, for such diversified graph scenarios;\\
  \textbf{\large{-}} \textbf{2.} \textbf{\emph{KD}} is resource-\emph{inefficient}. The distilled model from \emph{KD} only handles exactly the same task as the teacher does. In other words, every student is always unique to a single task, leading to model redundancy for multi-task scenarios. 

\noindent
\textbf{Proposed Novel Deep Graph Reprogramming (\textsc{Gare}).}
Driven by the challenges of the \emph{KD}-based model reusing scheme, we develop in this paper a novel paradigm of \emph{deep graph reprogramming} (\textsc{Gare}) for more generalized and resource-efficient model reusing, that explicitly considers the topological uniqueness of graph data:
\vspace{-0.6em}
\begin{task}[\textbf{Reusing GNNs via Deep Graph Reprogramming}]
  \label{task:2}
  \textit{Deep graph reprogramming aims to reuse a pre-trained model, without changing any architecture nor parameter, for a bunch of various-domain and cross-level downstream tasks, via reprogramming graph data or model behaviors.}
\end{task}
\vspace{-0.8em}
As such, the proposed \textsc{Gare} is ideally superior to the ubiquitous \emph{KD} with the following merits:\\
\textbf{\large{+}}
\textbf{1.}
\textbf{\textsc{Gare}} allows for the reuse of a \emph{single} pre-trained GNN for \emph{multiple} cross-level/domain downstream tasks and datasets, as shown in Fig.~\ref{fig:intro}, thereby getting rid of the \emph{KD} restriction  on well-provided pertinent pre-trained models;\\
\textbf{\large{+}}
\textbf{2.}
\textbf{\textsc{Gare}} is free of re-training or fine-tuning, unlike \emph{KD} that substantially re-trains a student model from scratch, thereby making it possible for deployment in resource-constrained environments such as edge computing;\\
\textbf{\large{+}}
\textbf{3.}
\textbf{\textsc{Gare}} is memory-efficient, where a pre-trained model with \textsc{Gare} is anticipated to be versatile and multi-talented that integrates the expertise of multiple tasks.

\subsection{Challenges Towards G\textsc{\textbf{ARE}}  }

The ambitious goal of \textsc{Gare} in Task~\ref{task:2} is primarily accomplished with the two key challenges:\\
\textbf{\Large{$\divideontimes$}} \textbf{Data Side:} The first issue to be tackled regards handling various-dimension downstream features, considering that the pre-trained GNN in \textsc{Gare} is frozen without auxiliary transforming layers nor fine-tuning. For example, every node in the \emph{Cora} citation network has 1433 input features, whereas that in \emph{Amazon Co-purchase} graphs has 767 ones;\\
\textbf{\Large{$\divideontimes$}} \textbf{Model Side:} The second challenge towards \textsc{Gare} lies in the insufficient model capacity under the per-GNN-multiple-task scenario of \textsc{Gare}, especially for cross-domain downstream tasks as shown in Fig.~\ref{fig:intro}.

To tackle the data and model dilemmas of \textsc{Gare}, we devise a couple of data and model reprogramming paradigms, respectively, as will be elaborated in the following sections.

\subsection{Reprogramming Paradigms for G\textsc{\textbf{ARE}}}
\label{sect:preanalysis}

\subsubsection{Data Reprogramming (D\textsc{\textbf{ARE}})}
\label{sect:dare}
\vspace{-1.2mm}

\noindent
\textbf{Rationale Behind \textsc{Dare}.}
To tackle the problem of diversified features on the input side, a na\"ive idea is rearranging graph representations to adapt varying-dimension downstream features to
accommodate to the pre-trained GNN.
As such, the challenge instead comes to be how to adapt the target downstream features.  

To address this challenge of feature adaption, we paradoxically resort to a special type of adversarial attack, termed as \emph{adversarial reprogramming attack}, as introduced in Sect.~\ref{sect:relatedwork}.
In essence, the adversarial reprogramming attack demonstrates a security vulnerability of CNNs, where an attacker can easily redirect a model with perturbations to perform the selected task without letting the model providers know, thereby leading to ethical concerns, such as repurposing housekeeping robots to criminal activities.

Our idea here is to flip the role of adversarial reprogramming attacks, by turning the attackers that mean to perturb the model usage, into guards that aim to repurpose a pre-trained GNN to perform the intended downstream tasks.

However, adversarial reprogramming attack is formerly merely studied in the CNN domain. 
As such, it remains unknown in the machine learning community whether GNNs are also vulnerable to adversarial reprogramming attacks, which is a \emph{prerequisite} for the success in applying the idea of adversarial reprogramming to \textsc{Dare}.

To this end, we as attackers perform in Tab.~\ref{tab:attack} an evasion attack on graph data, that tries to repurpose a pre-trained GNN designated for \emph{product category prediction} to the new tasks of \emph{molecule classification} and \emph{molecule property regression}, through simply \emph{adding} the generated adversarial perturbations to the raw node features \cite{sun2018adversarial}.
We employ here the datasets of \emph{AmazonCoBuy} \cite{mcauley2015image,wang2019deep}, \emph{ogbg-molbbbp} \cite{wu2018moleculenet}, and \emph{ogbg-molesol} \cite{wu2018moleculenet} as examples.
Surprisingly, with such a vanilla manner, the attacked pre-trained GNNs are extraordinarily competent to handle the unseen tasks, which have a significant domain gap with the former ones, leading to the observation as follows:
\vspace{-0.6em}
\begin{remark}[\textbf{Adversarial Reprogramming Attacks on Graph Data}]
  \label{proposition:1}
  \textit{Graph neural networks are susceptible to adversarial reprogramming attacks, where
   an adversarial perturbation on graph data can readily repurpose a graph neural network to perform a task chosen by the adversary, without notifying the model provider.}
\end{remark}
\vspace{-0.6em}
\noindent
\textbf{Motivations of Heter-\textsc{Dare}-\emph{MetaFP} and Homo-\textsc{Dare}-\emph{EdgSlim}+\emph{MetaGP} Methods.}
Now, we turn our role back from attackers to reputable citizens that would like to reuse a pre-trained GNN to alleviate the training efforts for downstream tasks.
Remark~\ref{proposition:1} thereby illustrates that:\\
$\blacktriangleright$
\textbf{1.}
It is technically feasible to convert adversarial reprogramming attack to effective \textsc{Dare} on graph data, which \textbf{motivates} us to devise a universal \emph{Meta-FeatPadding (MetaFP)} approach, upon adversarial node feature perturbations, for heterogeneous-\textsc{Dare} (Sect.~\ref{sect:metafeatpadding});\\
$\blacktriangleright$
\textbf{2.}
Except for node-level perturbations, other adversarial example types tailored for graph data should also be effective for \textsc{Dare}, such as edge-level perturbations and structure-level perturbations \cite{sun2018adversarial}, \textbf{motivating} us to develop a \emph{transductive Edge-Slimming (EdgSlim)} (Sect.~\ref{sect:edgeslimming}) and an \emph{inductive Meta-GraPadding (MetaGP)} (Sect.~\ref{sect:graphpadding}) approaches, respectively, for homogenous-\textsc{Dare}.

\newcommand{\liuhaothreec}{\fontsize{5pt}{\baselineskip}\selectfont}
\newcommand{\liuhaothreetwoc}{\fontsize{4.5pt}{\baselineskip}\selectfont}
\begin{table}[t]
  \caption{Results of adversarial reprogramming attacks on graphs.}
  \vspace{-6.5mm}
  \begin{center}
  \liuhaothreec
  \setlength\tabcolsep{1 pt}
  {\renewcommand{\arraystretch}{0.5}
  \begin{tabular}{l|cc|cc}
    \noalign{\hrule height 0.5pt}
    \bf{Roles} & \bf Model Provider & \bf Adversarial Attacker & \bf Model Provider & \bf Adversarial Attacker \\
    \bf{Datasets} & \texttt{Computers} & \texttt{ogbg-molbbbp} & \texttt{Photo} & \texttt{ogbg-molesol} \\
    \bf{Task Types} & \liuhaothreetwoc{Computer-Category Prediction} & \liuhaothreetwoc{Molecule Classification} &  \liuhaothreetwoc{Photo-Category Prediction} & \liuhaothreetwoc{Molecule Regression} \\
    \hline
    \bf{Before Attack} & \emph{Accuary}: 0.9485 & -- & \emph{Accuary}: 0.9561& -- \\
    \bf{After Attack}  & -- & \emph{ROC-AUC}:  \bf 0.6132 & -- & \emph{RMSE}: \bf 2.7479\\
    \hline
    \bf{Re-training} & -- & \emph{ROC-AUC}: 0.6709 & -- & \emph{RMSE}: 1.3000 \\
    \noalign{\hrule height 0.5pt}
  \end{tabular}}
  \end{center}
  \vspace{-7.5mm}
  \label{tab:attack}
\end{table}

\vspace{-1.5mm}
\subsubsection{Model Reprogramming (M\textsc{\textbf{ERE}})}
\label{sect:mere}
\vspace{-1.2mm}
\noindent\textbf{Rationale Behind \textsc{Mere}.}
Backed by the theory of adversarial reprogramming attacks (Remark~\ref{proposition:1}), in most cases, a pre-trained model equipped with \textsc{Dare} in Sect.~\ref{sect:dare} can already achieve encouraging results in tackling various downstream tasks.
Despite its gratifying performance, we empirically observe that the downstream performance by only using \textsc{Dare} is prone to a bottleneck especially for some tasks that have considerable domain gaps
with the pre-trained one, 
for example the pre-trained task on paper analysis and the other one on e-commerce prediction.

We conjecture that such a bottleneck is due to the frozen GNN parameters and architectures, leading to insufficient expressive capabilities in modeling cross-domain topological properties.
Motivated by the above observation, we further develop a \textsc{Mere} paradigm to strengthen the model capacities, acting as a complement to \textsc{Dare} under the scenarios of tremendous-domain-gap GNN reusing.

To this end, a vanilla possible solution for model enhancement will be resorting to dynamic networks \cite{han2021dynamic} that are well-studied in the CNN domain. 
Plenty of dynamic inference schemes that are designated for CNNs, in fact, are equally feasible to the non-Euclidean domain of GNNs, such as early exiting, layer skipping, and dynamic routing.
Moreover, almost all these dynamic strategies require no changes to original model parameters, thus readily acting as a specific implementation of \textsc{Mere}.

Nevertheless, instead of simply using CNN-based dynamic network schemes, we perform in this paper a pilot study of \textsc{Mere} by explicitly considering the most critical characteristic that is unique to GNNs, namely \emph{message aggregation},
and leaving the explorations of other dynamic paradigms in \textsc{Mere} for future works.

In the literature,
message aggregation schemes have already been identified as one of the most crucial components in graph analysis, both empirically and theoretically \cite{corso2020principal}.
However, the significance of aggregation behaviors in model reusing has not yet been explored, which is a precondition for the success of aggregation-based \textsc{Mere}.

\begin{wrapfigure}{r}{4.9cm}
  \vspace{-0.36cm}
  \hspace{-0.45cm}
  \raggedleft
  \includegraphics[width=0.3\textwidth]{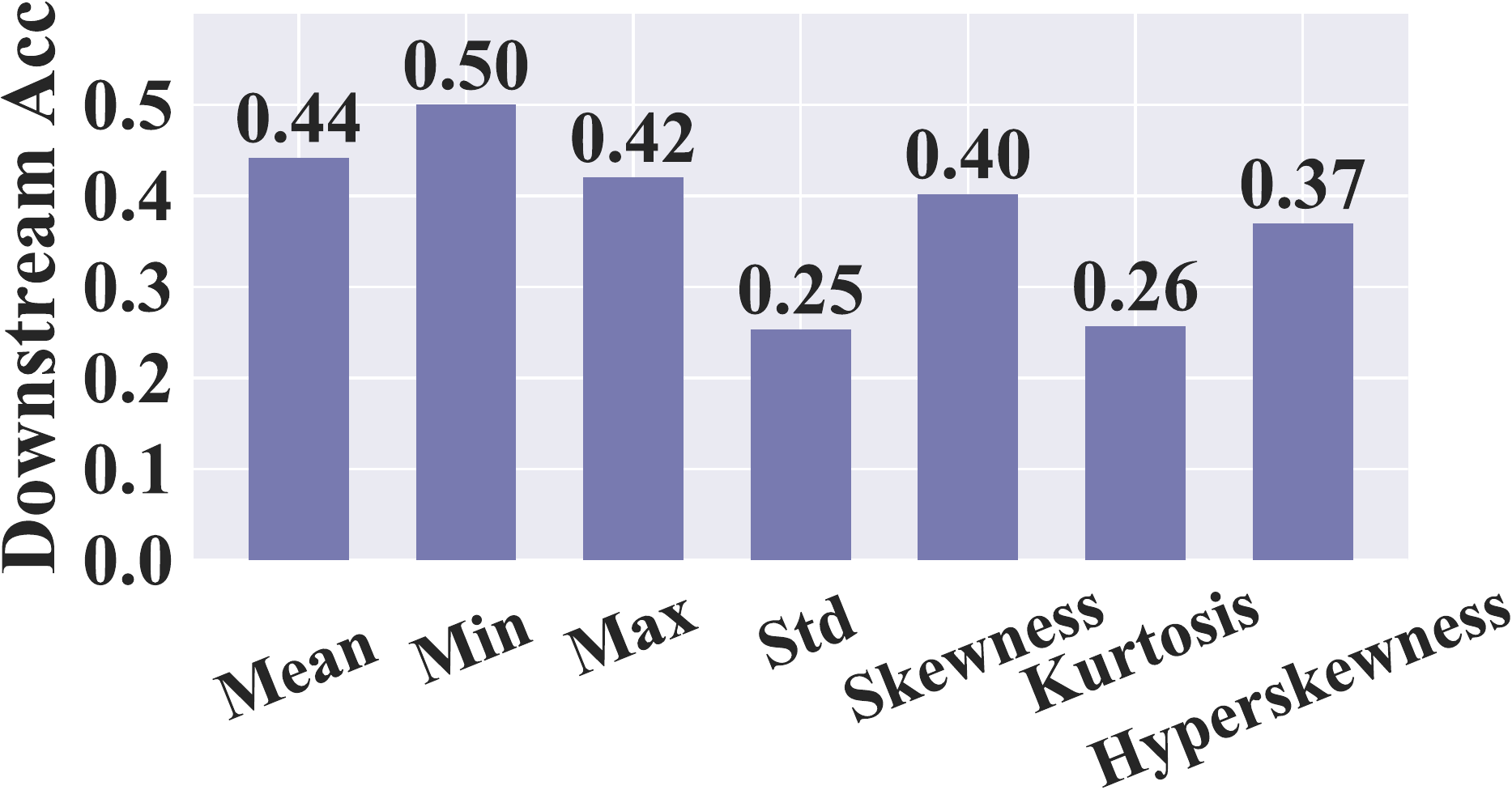}
  \vspace{-0.75cm}
  \captionsetup{font=footnotesize}
  \caption{Reusing with various aggregators.}
  \label{fig:bar}
  \vspace{-0.4cm}
  \end{wrapfigure}

To this end, we explore the \textsc{Mere} paradigm
by firstly performing a prior study with \emph{Cora} dataset in Fig.~\ref{fig:bar}, that attempts to observe the diverse performance of reusing a fixed GNN pre-trained for classifying the node categories of
{\small \{Case-Based, Genetic-Algorithm, Neural-Network, Probabilistic-Method\}}, 
to directly handle the separate downstream classes of {\small \{Reinforcement-Learning, Rule-Learning, Theory\}}, by only replacing the pre-trained aggregator with other aggregation methods.
Remarkably, different aggregators in Fig.~\ref{fig:bar} 
lead to distinct downstream performance,
which can be summarized as:
\vspace{-0.8em}
\begin{remark}[\textbf{Aggregation Matters for Reusing}]
  \label{proposition:2}
  \textit{Various aggregators lead to diversified downstream task performance with the same model. There exists an optimal aggregation method tailored for each pair of downstream tasks and pre-trained models.}
\end{remark}

\vspace{-1.33mm}
\noindent
$\blacktriangleright$ \textbf{Motivated} by Remark~\ref{proposition:2}, we accordingly derive a 
\emph{reprogrammable aggregating (ReAgg)} method as a specific implementation of the \textsc{Mere} paradigm, that aims to dynamically change the aggregation behaviors under various downstream scenarios, which will be elaborated in Sect.~\ref{sect:reagg}.

\section{Proposed Methods: Implementing D\textsc{\textbf{ARE}} and M\textsc{\textbf{ERE}} Paradigms}
\label{sect:method}

In this section, we instantiate the proposed paradigms of \textsc{Dare} (Sect.~\ref{sect:dare}) and \textsc{Mere} (Sect.~\ref{sect:mere}), by elaborating three \textsc{Dare} methods and one \textsc{Mere} approach, tailored for various scenarios of the \textsc{Gare}-based model reusing.

\subsection{Overview and Case Discussions}

As analyzed in Sect.~\ref{sect:preanalysis}, a vanilla method to achieve \textsc{Dare} is generating an adversarial feature perturbation as an addition to raw features.
However, such a na\"ive addition manner is prone to a heavy computational burden, especially for high-dimensional-feature scenarios, where we have to optimize 
an equally high-dimensional perturbation for downstream tasks.
Also, such perturbation addition manner completely changes all raw inputs, thereby intrinsically can be interpreted as transforming downstream data to pre-trained one for model reusing, leading to performance bottleneck when the data gap is too significant for transformation. 

Motivated by this observation, 
we resort to generating lower-dimensional perturbations as \emph{paddings} around raw features, \emph{never} amending any raw input feature. 
As such, the issues of both computational costs and troublesome transformation are simultaneously alleviated.

Despite its merits, such a perturbation padding manner can only be applicable to the scenario where pre-trained and downstream features have heterogeneous dimensions. 
Driven by this consideration, we propose to divide the \textsc{Gare} scenarios into \emph{three cases}, and explore them separately to devise the corresponding best-suited methods for more resource-efficient model reusing:\\
$\bullet$\
\textbf{Case \#1. Universal-Heter-\textsc{Dare}:} Heterogeneous dimensions between pre-trained and downstream features under both transductive and inductive settings, addressed by \emph{Meta-FeatPadding} in Sect.~\ref{sect:metafeatpadding};\\
$\bullet$\
\textbf{Case \#2. Transductive-Homo-\textsc{Dare}:} Homogenous pre-trained and downstream dimensions for transductive tasks, solved by \emph{Edge-Slimming} in Sect.~\ref{sect:edgeslimming};\\
$\bullet$\ 
\textbf{Case \#3. Inductive-Homo-\textsc{Dare}:} Homogenous input dimensions for inductive tasks, tackled by \emph{Meta-GraPadding} in Sect.~\ref{sect:graphpadding}.

Furthermore, we provide in Sect.~\ref{sect:reagg} an examplar implementation of the \textsc{Mere} paradigm, by proposing \emph{reprogrammable aggregation (ReAgg)} that aims to complement \textsc{Dare} on the model side for challenging downstream tasks.

\begin{figure}[t]
  \centering
  \includegraphics[width=0.475\textwidth]{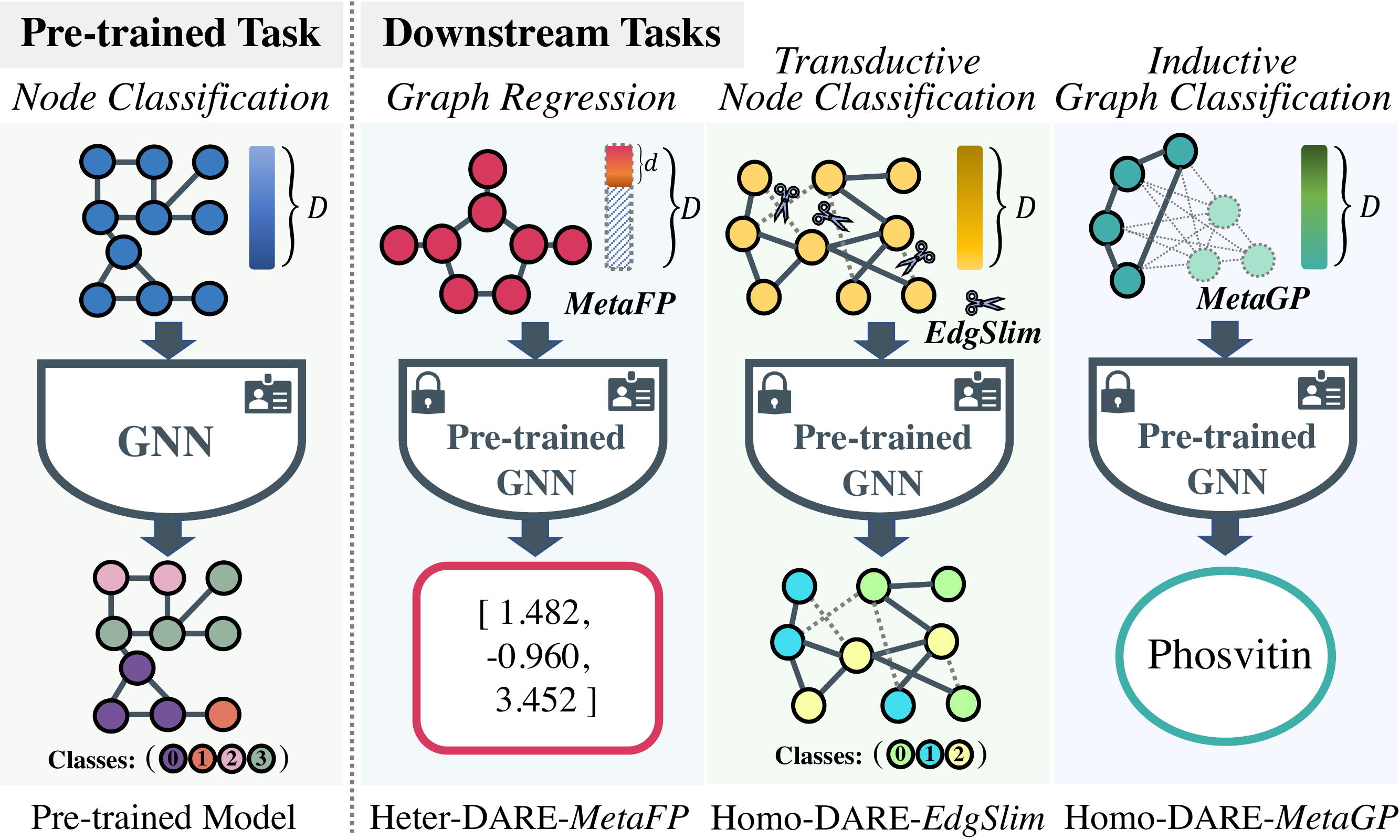}
  \vspace{-6mm}
  \caption{Illustration of the proposed approaches of \emph{MetaFP}, \emph{EdgSlim}, and \emph{MetaGP} for transductive and inductive \textsc{Dare} with heterogeneous and homogenous input dimensions.}
  \vspace{-2mm}
  \label{fig:method}
\end{figure}

\subsection{Universal Meta-FeatPadding for Heter-D\textsc{\textbf{ARE}}}
\label{sect:metafeatpadding}

The proposed \emph{Meta-FeatPadding (MetaFP)} aims to accommodate the diverse downstream feature dimensions, by padding around the raw features, supported by the theory of \emph{node-level} adversarial perturbations \cite{sun2018adversarial}.

Given a pre-trained model termed $\texttt{GNN}_\text{pre-trained}$, the process of generating the padded features for downstream tasks with \emph{MetaFP} can be formulated as follows:
\vspace{-1.5mm}
\begin{equation}
  \min _{\boldsymbol{\delta}_\textnormal{padding}} E_{(x, y) \sim \mathcal{D}}\left[ \mathcal{L}_{\text{downstream}}\left(\texttt{GNN}_\text{pre-trained}[x ||\boldsymbol{\delta}_\textnormal{padding}], y\right)\right],
  \label{eq:featpad}
\end{equation}
where $\mathcal{D}$ is the downstream data distribution, with 
$(x, y)$ denoting the downstream graph features and the associated labels, respectively.
Also, we use $||$ to represent the concatenation operation, which, in fact, performs feature padding that combines the optimized padding features $\boldsymbol{\delta}_\textnormal{padding}$ with the raw input features $x$.

As such, the task-specific $\boldsymbol{\delta}_\textnormal{padding}$ not only accommodates the feature dimensions of the downstream tasks to those of the pre-trained one with the frozen model of $\texttt{GNN}_\text{pre-trained}$, but also benefits the downstream performance by reducing the loss derived from the downstream loss function of $\mathcal{L}_{\text{downstream}}$.
The generation process of $\boldsymbol{\delta}_\textnormal{padding}$ is empirically very fast for most cases, where only one or several epochs are typically sufficient for converged results.

During inference, the optimized $\boldsymbol{\delta}_\textnormal{padding}$ on training downstream data is padded around all the testing downstream samples to obtain the prediction results. 
Furthermore,
for the case where the output downstream dimensions are not aligned with the pre-trained ones, we simply use the corresponding part of the pre-trained neurons at the final linear layer, which is a common avenue in dynamic networks.

\subsection{Transductive Edge-Slimming for Homo-D\textsc{\textbf{ARE}}}
\label{sect:edgeslimming}

The devised \emph{MetaFP} in Sect.~\ref{sect:metafeatpadding} is competent to tackle the heterogeneous-dimension case of \textsc{Gare}-based model reusing.
Despite its encouraging performance, \emph{MetaFP} is not effective in handling the downstream graph samples that have homogenous feature dimensions to the pre-trained ones, since it is no longer necessary to conduct meta padding for dimension accommodation.
As such, it remains challenging in such   homogenous-dimension cases, on performing \textsc{Dare} to adapt the pre-trained model to new tasks.

Driven by this challenge, we turn from \emph{node-level} perturbations to another type of adversarial graph examples, namely \emph{adversarial edge-level perturbations} \cite{sun2018adversarial}, that aim to attack the model by manipulating edges. 
Here, we flip again the attacker role of \emph{adversarial edge-level perturbations} to achieve resource-efficient model reusing, by modifying the node connections in the downstream graph data, meanwhile without changing raw node features, leading to the proposed \emph{Edge-Slimming (EdgSlim)} \textsc{Dare} approach.

To this end, we formulate the algorithmic process of \emph{EdgSlim} as a combinatorial optimization problem:
\vspace{-1.5mm}
\begin{equation}
  \begin{aligned}
\min_{\left\{u_i, v_i\right\}_{i=1}^{m}} & \sum_{i=1}^m\left|\frac{\partial \mathcal{L}_\text{downstream}}{\partial \alpha_{u_i, v_i}}\right| \\
\text { \emph{s.t.} } 
\tilde{\mathcal G}& =\operatorname{Modify}\left(\mathcal{G},\left\{\alpha_{u_i, v_i}\right\}_{i=1}^m\right) \\
& =\left(\mathcal{G} \ \backslash \left\{u_i, v_i\right\}\right), \ \text{if}\ \    \frac{\partial \mathcal{L}_\text{downstream}}{\partial \alpha_{u_i, v_i}}>0,
  \end{aligned}
  \label{eq:edgeslim}
  \vspace{-1mm}
\end{equation}
where $\{u, v\}$ denotes the connection between the node $u$ and $v$, with $m$ representing the total number of edges in the input graph $\mathcal{G}$.
$\mathcal{L}_\text{downstream}$ is the loss function for the downstream task. 
$\alpha_{u_i, v_i}$ is our constructed unary edge feature, such that we can compute the derivative of $\mathcal{L}_\text{downstream}$ with respect to the adjacency matrix of $\mathcal{G}$, with $\alpha_{u, v}=\mathbb{I}(u \in \mathcal{N}(v))$ where $\mathcal{N}(v)$ denotes the set of neighbors for the node $v$.
$\backslash$ represents the edge deletion operation.

As such, Eq.~\ref{eq:edgeslim} indicates that the proposed 
\emph{EdgSlim} sequentially slims the connections in the downstream graph of which the corresponding edge gradients are greater than $0$, starting from the edge with the largest gradients.   
The downstream loss can thereby be reduced by simply optimizing the connections.
Notably, similar to \emph{MetaFP}, the optimization with \emph{EdgSlim} converges very fast, typically with only several epochs, and meanwhile occupies limited resources.

\subsection{Inductive Meta-GraPadding for Homo-D\textsc{\textbf{ARE}}}
\label{sect:graphpadding}

In spite of the gratifying results of \emph{EdgSlim}, 
Eq.~\ref{eq:edgeslim} is not applicable to the inductive task setting, where plenty of graphs are received as inputs.
In this case, the edge slimming operation can only be performed on training graphs, not capable of transferring to the testing ones.
To alleviate this dilemma, we propose a \emph{Meta-GraPadding (MetaGP)} method to tackle the inductive \textsc{Gare} scenarios, where the downstream features have the same dimensions as the pre-trained ones, as illustrated in Fig.~\ref{fig:method}.

Our design of \emph{MetaGP} is driven by the \emph{structure-level perturbation} in adversarial examples \cite{sun2018adversarial}.
In particular,
instead of padding the generated perturbations around raw node features, the proposed \emph{MetaGP} yields a tiny subgraph, with only, for example, ten nodes, which is then padded around every downstream graph, of which each meta node connects the downstream graph nodes in a fully-connected manner.
The features in the introduced meta graph are generated in the same way as that of yielding padded features in Eq.~\ref{eq:featpad}.
At the inference stage, the learned meta-graph is padded around all the testing graphs, leading the pre-trained GNN to perform the target downstream inductive task.

The process of generating the meta-graph in \emph{MetaGP} is computation-efficient, where a meta-graph with only ten nodes is typically sufficient for most tasks.
Moreover, the feature generation procedure is also lightweight, 
given the property of inductive graph learning tasks where the input features are generally low-dimension, \eg,
the \emph{QM7b} dataset having only 1-dimension features, as well as
the \emph{ogbg-molbace},
\emph{ogbg-molbbbp}, and
\emph{ogbg-molesol} datasets with an input feature dimension of nine.

\begin{table*}[!t]
  \caption{Results of reusing a pre-trained model on {\footnotesize \emph{Citeseer}} to simultaneously handle four unseen tasks with heterogeneous dimensions and objectives, averaged with 20 independent runs. 
  ``Re-training'' indicates whether the pre-trained parameters are changed. 
  Notably, the \nth{8} line shows that \emph{ReAgg} is more competent for the large-domain-gap scenarios (2.3\% improvement averagely),
  but slightly falls behind for \emph{similar}-domain tasks, such as {\footnotesize \{\emph{Cora}, \emph{Citeseer}\}}, both of which classify computer science papers.
  Also, our \emph{MetaFP} yields stable results that only slightly vary with padding initializations, with standard deviations of {\footnotesize \{0.0030, 0.0023, 0.0006, 0.0008\}} for the four downstream tasks.}
  \vspace{-6mm}
  \begin{center}
  \scriptsize
  \setlength\tabcolsep{3.4 pt}
  {\renewcommand{\arraystretch}{0.95}
  \begin{tabular}{l|c|c|c|ccccc}
    \noalign{\hrule height 0.8pt}
    \multirow{2}{*}{\textbf{Methods}} & \multirow{1}{*}{\textbf{Model}} & \multirow{2}{*}{\textbf{Model Parameter Sizes}} & \multicolumn{1}{c|}{\textbf{Pre-trained Task}} & \multicolumn{4}{c}{\textbf{Downstream Heterogeneous Tasks}}\\ 
    & \multirow{1}{*}{\textbf{Re-training?}} & & {{\texttt{Citeseer}}} & {{\texttt{Cora}}}& {{\texttt{Pubmed}}}& {{\texttt{Computers}}}& {{\texttt{Photo}}}\\

    \noalign{\hrule height 0.5pt}
    Pre-trained Model \cite{velivckovic2018graph} & \footnotesize{$\times$} &474.89K & 0.7950 & N/A & N/A & N/A & N/A\\
    \hline
    Training from Scratch \cite{velivckovic2018graph} & \footnotesize{$\surd$} & \{474.89K, 184.33K, 64.52K, 99.09K, 96.27K\} & 0.7950 & 0.9144 & 0.8530& 0.9475&0.9555 \\
    Reusing via Fine-tuning \cite{hu2019strategies} & \footnotesize{$\surd$} & 474.89K & 0.7950 &0.8710 &0.8860&0.9542 &0.9555  \\
    Multi-task Learning \cite{capela2019multitask} + SlimGNN \cite{jing2021amalgamating} & \footnotesize{$\surd$} 
    &477.62K & 0.7880 &0.8780 & 0.8450 & 0.9108 & 0.9317 \\
    Vanilla Reusing \cite{velivckovic2018graph} + SlimGNN \cite{jing2021amalgamating} & \footnotesize{$\times$} &474.89K & 0.7950 &0.1571&0.3250 & 0.5037 &0.2183 \\
    
    \bf Ours (\emph{MetaFP}) & \bf \footnotesize{$\boldsymbol\times$} & \bf 474.89K & \bf 0.7950 &\bf 0.8335 &\bf 0.7790 &\bf 0.9085& \bf 0.8909\\
    \bf Ours (\emph{MetaFP} + \emph{ReAgg}) & \bf\footnotesize{$\boldsymbol\times$} &\bf 474.89K &\bf 0.7950 &\bf 0.8312 & \bf 0.8030&\bf 0.9229& \bf 0.9213 \\
    \noalign{\hrule height 0.8pt}
  \end{tabular}}
  \end{center}
  \label{tab:hetermain}
  \vspace{-6mm}
  \end{table*}

\subsection{Reprogrammable Aggregating for M\textsc{\textbf{ERE}}}
\label{sect:reagg}

With the three elaborated \textsc{Dare} methods demonstrated in the preceding sections, a pre-trained GNN can already achieve empirically encouraging results in various downstream tasks and settings.
To further improve the reusing performance especially under the 
large-domain-gap scenarios,
we propose here a \emph{reprogrammable aggregating (ReAgg)} method as a pilot study of the \textsc{Mere} paradigm. 

Driven by 
Remark~\ref{proposition:2}, the goal of the proposed \emph{ReAgg} is to adaptively determine the optimal aggregation behaviors conditioned on different downstream tasks, without changing model parameters, thereby strengthening the model capacities. 
However, such an ambitious goal comes with the challenge of the undifferentiable discrete decisions of aggregators. 
To address this challenge, one possible solution is resorting to \emph{reinforcement learning (RL)}. 
However, it is a known issue that \emph{RL} is prone to a high computation burden, due to its Monte Carlo search process. 
Another solution is to use the \emph{improved SemHash} technique \cite{kaiser2018discrete} for discrete optimization. 
However, we empirically observe that \emph{improved SemHash} for \textsc{Mere} is likely to cause the collapse issue, where a specific aggregator is always or never picked up.

Motivated by the above analysis, we propose to leverage Gumbel-Max trick \cite{veit2020convolutional} for \emph{ReAgg}, which is a more 
prevalent strategy for optimizing discrete variables than \emph{improved SemHash} in dynamic neural networks \cite{han2021dynamic}.
In particular,
to alleviate the dilemma of model collapse, 
we propose incorporating stochasticity into the aggregator decision process with the well-studied Gumbel sampling \cite{maddison2014sampling,veit2020convolutional}.
We then propagate the gradients via the continuous form of the Gumbel-Max trick \cite{jang2016categorical}.
Specifically,
despite the capability in parameterizing discrete distributions, the Gumbel-Max trick is, in fact, dependent on the \emph{argmax} operation, which is non-differentiable.
To address this issue, we thereby employ its continuous relaxation form of the Gumbel-softmax estimator that replaces \emph{argmax} with a \emph{softmax} function.

The detailed process of determining the optimal aggregation manner for each downstream task can be formulated as:
$	\text{Aggregator}_\textbf{k} = \text{softmax}\big((\mathcal{F}(\mathcal{G}) + G) / \tau \big)$, where \textbf{k} denotes the \emph{k}-th downstream task.
Also, $G$ denotes the sampled Gumbel random noise, which introduces stochasticity to avoid the collapse problem.
$\mathcal{F}$ represents the intermediate features with $\mathcal{G}$ as inputs.
$\tau$ is a constant denoting the softmax temperature.
We clarify that for superior performance, $\mathcal{F}$ can be generated by feeding $\mathcal{G}$ into a transformation layer, which lies out of the pre-trained model and does not directly participate in the inference process as a part of the Gumbel-softmax estimator.
In this way, the proposed \emph{ReAgg} adaptively determines the optimal aggregator conditioned on each task, also without changing any pre-trained parameters, thereby enhancing the model capability.

\section{Experiments}
\label{sec:exp}

We evaluate the performance of a series of \textsc{Dare} and \textsc{Mere} approaches on fourteen publicly available benchmarks.
Here we clarify that our goal in the experiments is \emph{not} to achieve the state-of-the-art performance, but rather reusing a pre-trained GNN to yield favorable results for as many downstream tasks as possible under limited computational resources.   

\newcommand{\liuhaotwo}{\fontsize{6pt}{\baselineskip}\selectfont}
\begin{table}[t]
  \vspace{-1mm}
  \caption{Ablation studies of diverse padding sizes/positions and various pre-trained/downstream tasks. Notably, our \emph{MetaFP} is effective even with tiny sizes and random positions.}
  \vspace{-6mm}
  \begin{center}
  \scriptsize
  \setlength\tabcolsep{3.6 pt}
  {\renewcommand{\arraystretch}{1.1}
  \begin{tabular}{l|c|cccc}
    \noalign{\hrule height 0.8pt}

    \textbf{Set of Heterogeneous Tasks}& \textbf{Padding} & \multicolumn{4}{c}{\textbf{Padding Positions}}\\ 
  \multirow{1}{*}{\textbf{\{{{Pre-trained}}, {{Downstream}}\}}} & \textbf{Size} & Front & Center & End & Random \\
    \noalign{\hrule height 0.5pt}
    {\{{\texttt{Computers}}, {\texttt{Photo}}\}} & 22 & 0.9183 & 0.9161 & 0.9212 & 0.9165 \\
    {\{{\texttt{Photo}}, {\texttt{Pubmed}}\}}& 245 & 0.8420 & 0.8430 & 0.8420 & 0.8440\\
    {\{{\texttt{Computers}}, {\texttt{Pubmed}}\}}& 267 & 0.8300 & 0.8320 &0.8370  & 0.8330\\
    {\{{\texttt{Cora}}, {\texttt{Computers}}\}}& 666 & 0.8585 & 0.8792 & 0.8561 & 0.8910\\
    {\{{\texttt{Cora}}, {\texttt{Photo}}\}}& 688 & 0.8337 & 0.8785 & 0.8402 & 0.8915\\
    {\{{\texttt{Cora}}, {\texttt{Pubmed}}\}}& 933 & 0.8180 & 0.8210 & 0.8200 & 0.8240 \\
    {\{{\texttt{Citeseer}}, {\texttt{Cora}}\}}& 2270 & 0.8370 & 0.8335 & 0.8417 & 0.8535 \\
    {\{{\texttt{Citeseer}}, {\texttt{Pubmed}}\}}& 3203 & 0.7790 & 0.7750 & 0.7740  & 0.8010\\
    \noalign{\hrule height 0.8pt}
  \end{tabular}}
  \end{center}
  \label{tab:padsize}
  \vspace{-6mm}
  \end{table}

\subsection{Experimental Settings}

\vspace{-0.5mm}
\noindent\textbf{Implementation Details.} 
Detailed dataset statistics can be found in the supplement.
In particular, we follow \cite{hu2020open,jing2021amalgamating} to split {\small\emph{Amazon Computers}}, {\small\emph{Amazon Photo}}, and the OGB datasets, whereas for {\small\emph{Cora}}, {\small\emph{Citeseer}}, and {\small\emph{Pubmed}} datasets, we use the splitting protocol in the supervised scenario for more stable results, as also done in \cite{chen2018fastgcn}.
Task-by-task architectures
and hyperparameter settings 
can be found in the supplementary material.
All the experiments are performed using a single NVIDIA GeForce RTX 2080 Ti GPU.

\noindent\textbf{Comparison Methods.}
Given our novel \textsc{Gare} setting,
there are few existing methods in the literature for a fair comparison, either with distinct task settings or inconsistent objectives.
As such, we derive two possible solutions that partly match our task.
Specifically, we derive a \emph{reusing via fine-tuning} approach that reuses a GNN via fine-tuning the parameters for downstream tasks.
Moreover, a \emph{multi-task-learning+SlimGNN} pipeline is also developed, that trains a slimmable graph convolution \cite{jing2021amalgamating} from scratch to accommodate the pre-trained and downstream feature dimensions.

\begin{figure}[t]
  \vspace{-0.5mm}
  \setlength\tabcolsep{1.5 pt}
  {\renewcommand{\arraystretch}{0.5}
  \begin{tabular}{cccc}
  \centering
  \includegraphics[height=2.12cm]{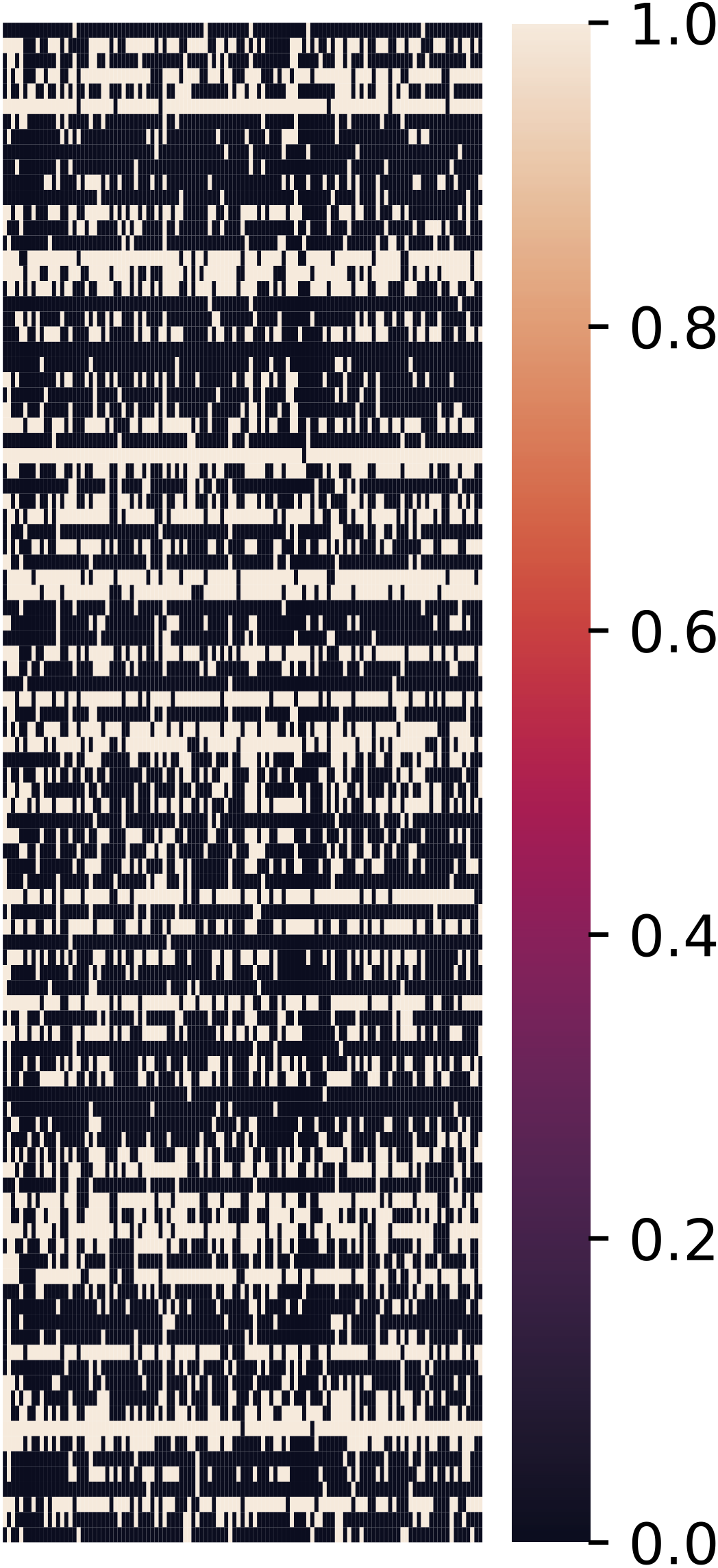}&\includegraphics[height=2.12cm]{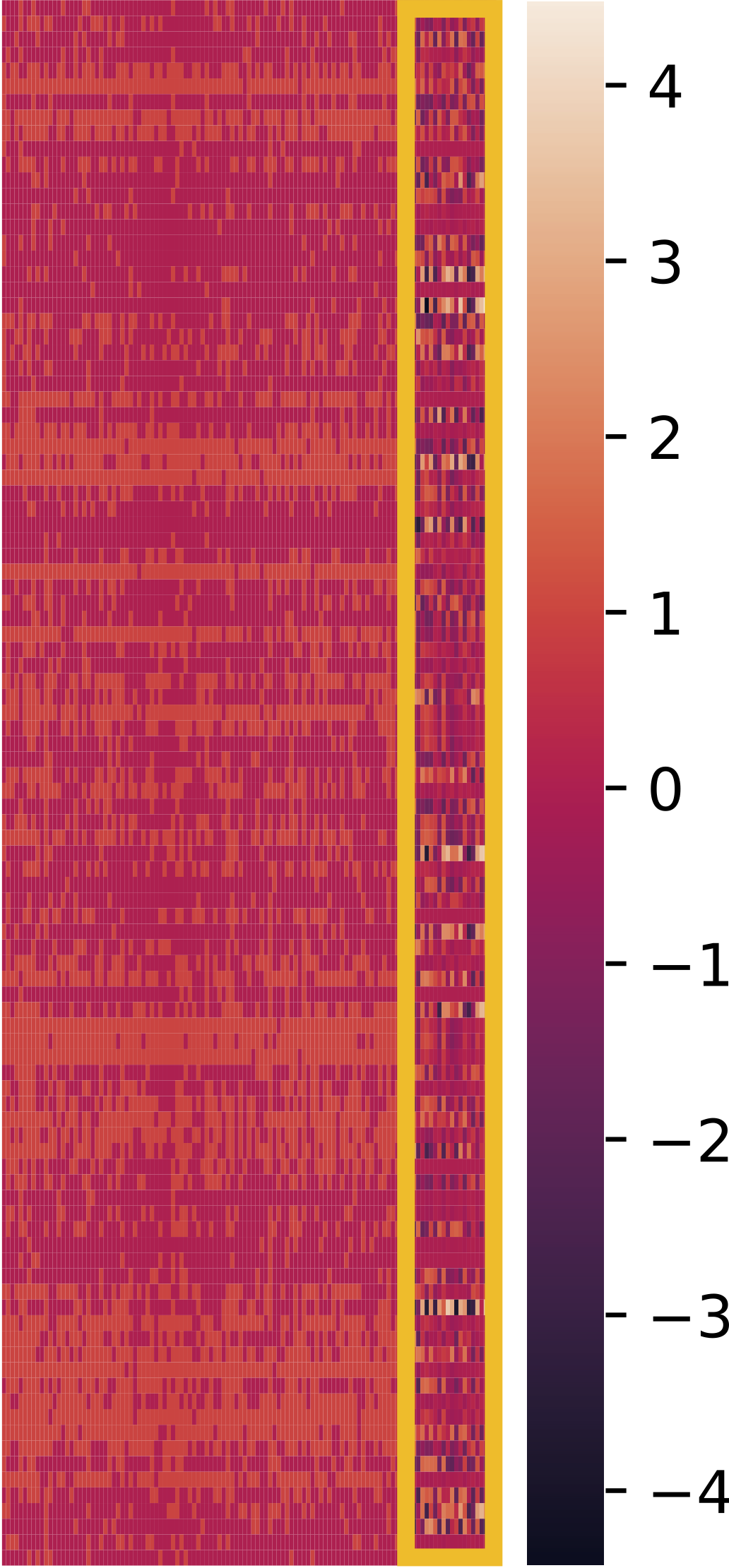}& \includegraphics[height=2.12cm]{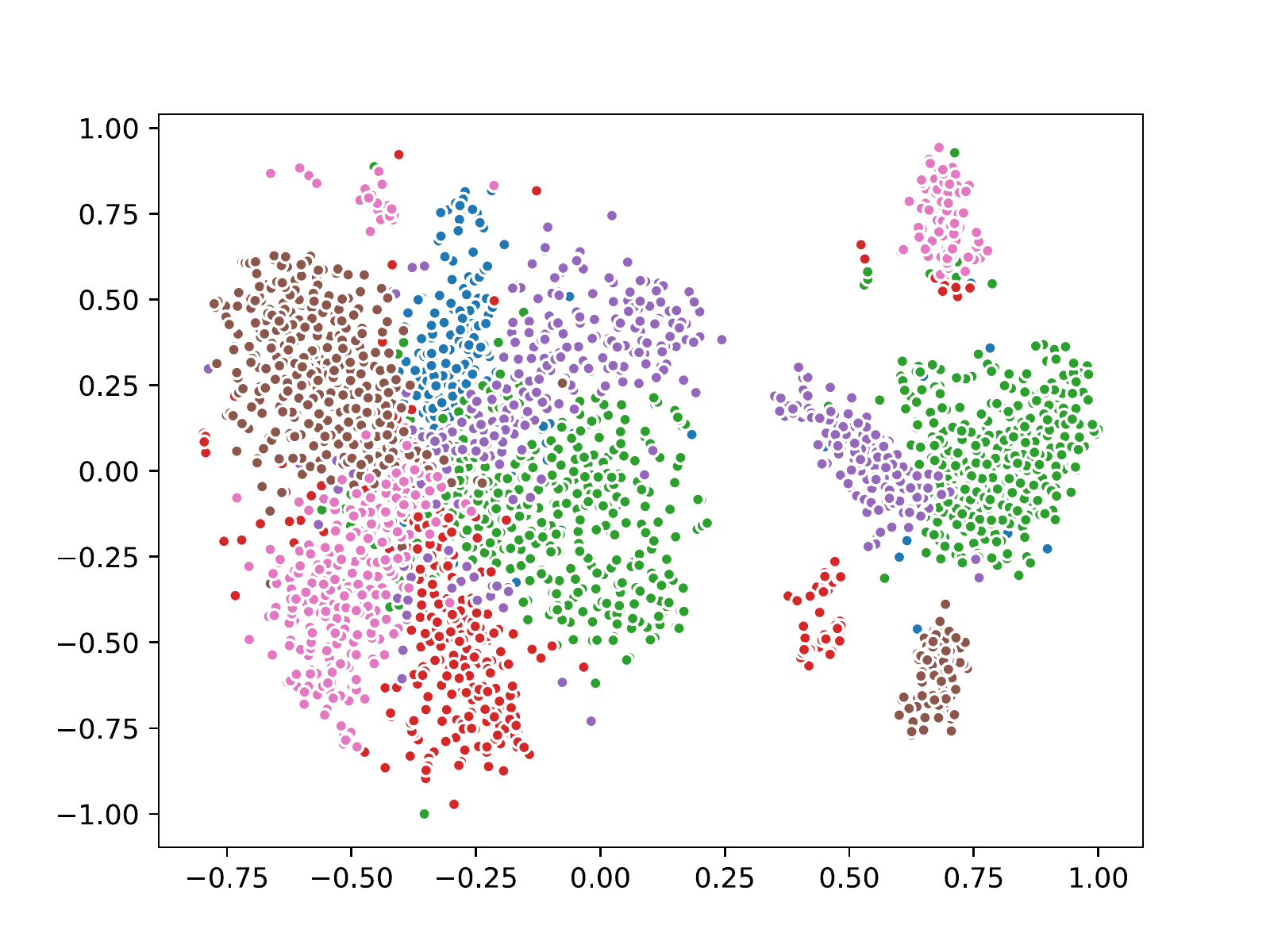}  & \includegraphics[height=2.12cm]{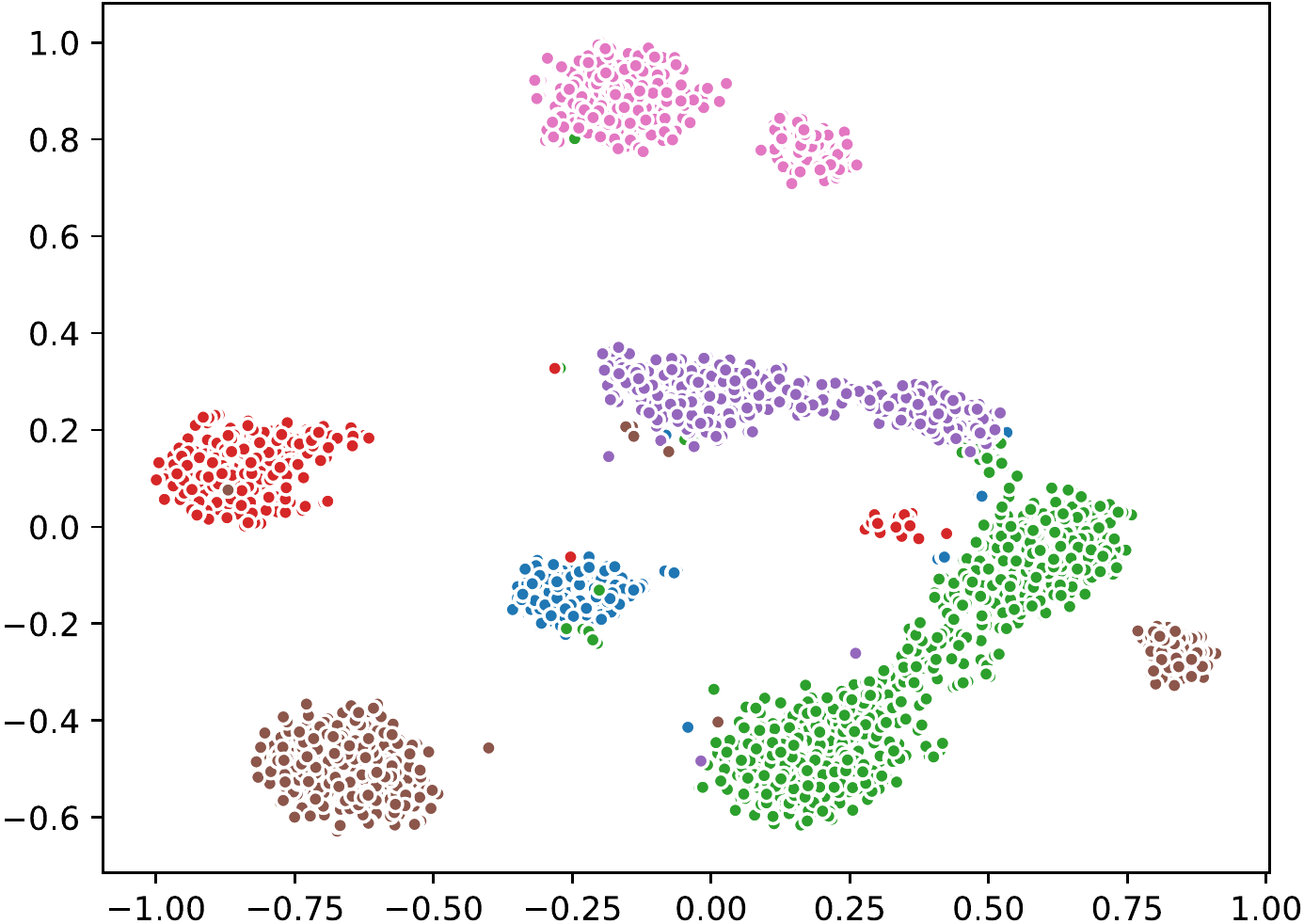}\\
  \smallskip
  \footnotesize{(a)} & \footnotesize{(b)} & \footnotesize{(c)} & \footnotesize{(d)} \\
  \end{tabular}
  }
  \vspace{-4.5mm}
    \caption{Feature/t-SNE visualizations of (a, c) before padding and (b, d) after padding,
    with the yellow frame indicating the paddings.
    }
    \label{fig:tsne} 
  \vspace{-5mm}
  \end{figure}

\newcommand{\liuhaofour}{\fontsize{7pt}{\baselineskip}\selectfont}
\begin{table}[b]
  \vspace{-4mm}
  \caption{Results of reusing a single model of a node-level task to directly tackle graph regression and graph classification tasks.
  }
  \vspace{-6mm}
  \begin{center}
  \scriptsize
  \setlength\tabcolsep{1.6 pt}
  {\renewcommand{\arraystretch}{1.1}
  \begin{tabular}{l|cc|cc}
    \noalign{\hrule height 0.8pt}
    \textbf{Pre-trained} Task & \multicolumn{4}{c}{{\texttt{Photo}}}\\ 
    {\textbf{Heterogeneous} Task Type} & \multicolumn{4}{c}{\bf\emph{Product Category Prediction}} \\ \noalign{\hrule height 0.2pt}
    {Pre-trained Results} & \multicolumn{4}{c}{{\emph{Acc}:  0.9561 } } \\
    \noalign{\hrule height 0.6pt}
    \textbf{Downstream} Tasks & \multicolumn{2}{c|}{{\texttt{QM7b}}} & \multicolumn{2}{c}{\texttt{PROTEINS}} \\
    {\textbf{Heterogeneous} Task Types} & \multicolumn{2}{c|}{\bf\emph{Molecule Regression}} & \multicolumn{2}{c}{\bf\emph{Protein Prediction}} \\ \noalign{\hrule height 0.2pt}
    Reusing Methods & \multicolumn{1}{c!{\vrule width 0.1pt}}{{Vanilla}} & \multicolumn{1}{c|}{{Ours}}  & \multicolumn{1}{c!{\vrule width 0.1pt}}{{Vanilla}} & \multicolumn{1}{c}{{Ours}} \\ \noalign{\hrule height 0.1pt}

    Downstream Results & \multicolumn{1}{c!{\vrule width 0.1pt}}{\emph{\liuhaofour MAE}\liuhaofour: 24.18} & \multicolumn{1}{c|}{\emph{\liuhaofour MAE}\liuhaofour :  \textbf{2.3093}} &\multicolumn{1}{c!{\vrule width 0.1pt}}{\emph{\liuhaofour Acc}\liuhaofour: 0.3304 } &\multicolumn{1}{c}{\emph{\liuhaofour Acc}\liuhaofour: \textbf{0.6071}} \\ \noalign{\hrule height 0.8pt}
    Re-training from Scratch & \multicolumn{2}{c|}{\emph{\liuhaofour MAE}\liuhaofour: 0.7264} & \multicolumn{2}{c}{\emph{\liuhaofour Acc}\liuhaofour: 0.6964} \\ 

    \noalign{\hrule height 0.8pt}
  \end{tabular}
  \begin{tabular}{l|cc|cc}
    \noalign{\hrule height 0.8pt}
  
    \textbf{Pre-trained} Task & \multicolumn{4}{c}{{\texttt{Cora}}}\\ 
    {\textbf{Heterogeneous} Task Type} & \multicolumn{4}{c}{\bf\emph{Publication Classification}} \\ \noalign{\hrule height 0.2pt}
    {Pre-trained Results} & \multicolumn{4}{c}{{\emph{Acc}: 0.9121} } \\
    \noalign{\hrule height 0.6pt}
    \textbf{Downstream} Tasks & \multicolumn{2}{c|}{{\texttt{QM7b}}} & \multicolumn{2}{c}{\texttt{PROTEINS}} \\
    {\textbf{Heterogeneous} Task Types} & \multicolumn{2}{c|}{\bf\emph{Molecule Regression}} & \multicolumn{2}{c}{\bf\emph{Protein Prediction}} \\ \noalign{\hrule height 0.2pt}
    Reusing Methods & \multicolumn{1}{c!{\vrule width 0.1pt}}{{Vanilla}} & \multicolumn{1}{c|}{{Ours}}  & \multicolumn{1}{c!{\vrule width 0.1pt}}{{Vanilla}} & \multicolumn{1}{c}{{Ours}} \\ \noalign{\hrule height 0.1pt}

    Downstream Results & \multicolumn{1}{c!{\vrule width 0.1pt}}{\emph{\liuhaofour MAE}\liuhaofour : 13.04} & \multicolumn{1}{c|}{\emph{\liuhaofour MAE}\liuhaofour : \textbf{0.8889}} &\multicolumn{1}{c!{\vrule width 0.1pt}}{\emph{\liuhaofour Acc}\liuhaofour : 0.4018} &\multicolumn{1}{c}{\emph{\liuhaofour Acc}\liuhaofour : \textbf{0.5893}} \\ \noalign{\hrule height 0.8pt}
    Re-training from Scratch & \multicolumn{2}{c|}{\emph{\liuhaofour MAE}\liuhaofour : 0.7264} & \multicolumn{2}{c}{\emph{\liuhaofour Acc}\liuhaofour : 0.6964 } \\ 

    \noalign{\hrule height 0.8pt}
  \end{tabular}}
  \end{center}
  \label{tab:hetergraph}
  \vspace{-4mm}
  \end{table}

\subsection{Reprogramming in Heterogeneous Domains}

\begin{wrapfigure}{r}{3.88cm}
  \vspace{-0.48cm}
  \hspace{-2cm}
  \raggedleft
  \includegraphics[width=0.24\textwidth]{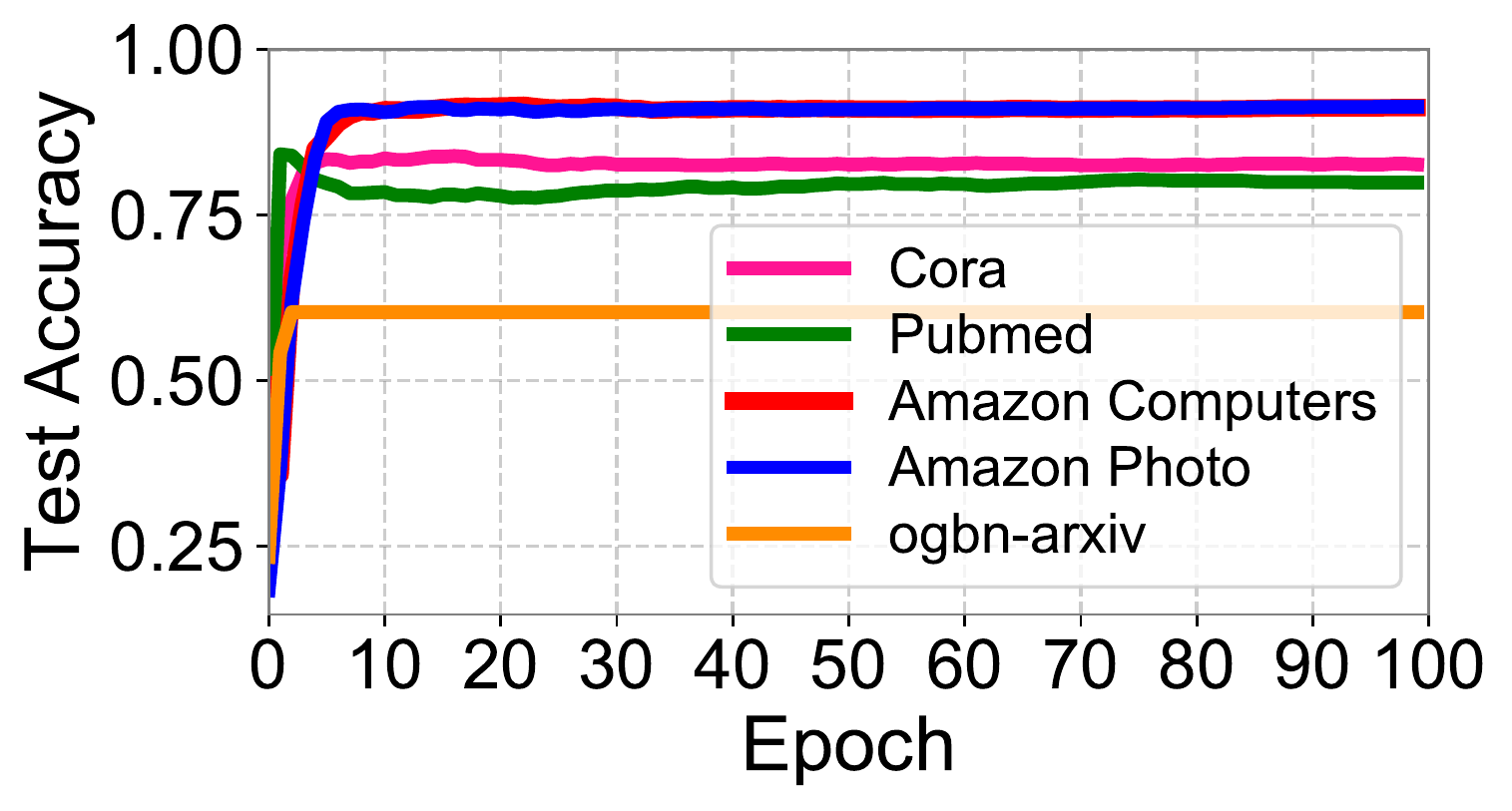}
  \vspace{-0.4cm}
  \captionsetup{font=footnotesize}
  \caption{Convergence speed.}
  \label{fig:convergence}
  \vspace{-0.4cm}
  \end{wrapfigure}

\vspace{-0.6mm}
\noindent\textbf{Heterogeneous Node Property Prediction.} 
We show in Tab.~\ref{tab:hetermain} the results of reprogramming a pre-trained GNN for a bunch of cross-domain node classification tasks, and further give in Tab.~\ref{tab:padsize} the ablation studies of diverse padding sizes and positions as well as different pre-trained and downstream tasks.
The \nth{6} line of Tab.~\ref{tab:hetermain} shows that \emph{MetaFP} makes it possible for cross-domain GNN reusing. 
Also,
the proposed \emph{ReAgg} for \textsc{Mere} further improves the downstream performance by about 2.3\% on average (the \nth{7} line of Tab.~\ref{tab:hetermain}).
Moreover, the visualization results before and after applying \emph{MetaFP} are demonstrated in Fig.~\ref{fig:tsne}.
Furthermore, we'd also like to highlight in Fig.~\ref{fig:convergence} that our method reaches convergence with only a few epochs, making it possible for deployment in resource-constrained environments.

\noindent\textbf{Heterogeneous Graph Classification and Regression.} 
Tab.~\ref{tab:hetergraph} shows the results of reusing a GNN for the more challenging cross-level tasks, indicating 
our proficiency in such cross-level scenarios.

\subsection{Reprogramming in Homogenous Domains}

\noindent\textbf{Homogenous Node Property Prediction.}
We perform in Tab.~\ref{tab:homoarxiv} extensive experiments of reusing a GNN for homogenous downstream tasks in a class-incremental setting.
The proposed \emph{EdgSlim}, as shown in Tab.~\ref{tab:homoarxiv}, achieves competitive performance at a low computational cost (Fig.~\ref{fig:convergence}).

\begin{table}[H]
  \vspace{-2mm}
  \caption{Results of Homo-\textsc{Dare} that adapts a pre-trained node property prediction model ({\emph{ogbn-arxiv-s1}}) to handle \emph{20} unseen \emph{homogenous} categories ({\emph{ogbn-arxiv-s2}}) in ogbn-arxiv dataset \cite{hu2020open}.
  }
  \vspace{-6.5mm}
  \begin{center}
  \scriptsize
  \setlength\tabcolsep{0.45 pt}
  {\renewcommand{\arraystretch}{1.05}
  \begin{tabular}{l|c|c|c|c}
    \noalign{\hrule height 0.8pt}
  
    \multirow{2}{*}{\textbf{Homogenous Multi-class}} & \multirow{2}{*}{\textbf{\shortstack{Re-\\train?}}} & \multirow{2}{*}{\textbf{Params}} & \multicolumn{1}{c|}{\textbf{Pre-trained Task}} & \multicolumn{1}{c}{\textbf{Downstream Task}}\\ 
    & & & {\texttt{ogbn-arxiv-s1}} & {\texttt{ogbn-arxiv-s2}}\\
    \noalign{\hrule height 0.1pt}
    \textbf{Number of Classes} & - & - & 20 & 20 \\
    \noalign{\hrule height 0.5pt}
    Pre-trained Model \cite{velivckovic2018graph} & \footnotesize{$\times$} & 35.75K & 0.7884 & N/A \\
    \hline
    Training from Scratch \cite{velivckovic2018graph} & \footnotesize{$\surd$} & 35.75K & N/A & 0.8115 \\
    Reusing via Fine-tuning \cite{hu2019strategies} & \footnotesize{$\surd$} & 35.75K & N/A & 0.8112  \\
    Multi-task Learning \cite{capela2019multitask} & \footnotesize{$\surd$} & 38.35K & 0.6387 & 0.6776 \\
    Vanilla Reusing \cite{velivckovic2018graph} & \footnotesize{$\times$} & 35.75K & N/A & 0.2334 \\

    \bf Ours (\emph{EdgSlim}) & \bf \footnotesize{$\boldsymbol\times$} & \bf 35.75K & \bf 0.7884 & \bf 0.6034 \\
    \noalign{\hrule height 0.8pt}
  \end{tabular}}
  \end{center}
  \label{tab:homoarxiv}
  \vspace{-5.6mm}
  \end{table}

\noindent\textbf{Homogenous Graph Classification and Regression.} We show in 
Tab.~\ref{tab:homograph} the cross-domain results for homogenous graph-level tasks, where our \emph{MetaGP} is proficient in reusing a graph classification model for the task of graph regression.

\newcommand{\liuhaothree}{\fontsize{6pt}{\baselineskip}\selectfont}

\begin{table}[H]
  \vspace{-2mm}
  \caption{Results of homogenous cross-domain graph-level tasks. 
  }
  \vspace{-6.5mm}
  \begin{center}
  \scriptsize
  \setlength\tabcolsep{0.9 pt}
  {\renewcommand{\arraystretch}{1.1}
  \begin{tabular}{l|cc|cc}
    \noalign{\hrule height 0.8pt}
  
    \textbf{Pre-trained} Task & \multicolumn{4}{c}{\texttt{ogbg-molbace}}\\ 
    {\textbf{Homogenous} Task Type} & \multicolumn{4}{c}{\bf\emph{Molecular Classification}} \\ \noalign{\hrule height 0.2pt}
    {Pre-trained Results} & \multicolumn{4}{c}{{\emph{ROC-AUC}:} 0.7734} \\
    \noalign{\hrule height 0.6pt}
    \textbf{Downstream} Tasks & \multicolumn{2}{c|}{\texttt{ogbg-molbbbp}} & \multicolumn{2}{c}{\texttt{ogbg-molesol}} \\
    {\textbf{Homogenous} Task Types} & \multicolumn{2}{c|}{\bf\emph{Molecular Classification}} & \multicolumn{2}{c}{\bf\emph{Molecular Regression}} \\ \noalign{\hrule height 0.2pt}
    Reusing Methods & \multicolumn{1}{c!{\vrule width 0.1pt}}{{Vanilla}} & \multicolumn{1}{c|}{{Ours}}  & \multicolumn{1}{c!{\vrule width 0.1pt}}{{Vanilla}} & \multicolumn{1}{c}{{Ours}} \\ \noalign{\hrule height 0.1pt}
    Downstream Results & \multicolumn{1}{c!{\vrule width 0.1pt}}{\emph{\liuhaothree{ROC-AUC}}\liuhaothree{: 0.5136}} & \multicolumn{1}{c|}{\emph{\liuhaothree{ROC-AUC}}\liuhaothree{: \textbf{0.6691}}} &\multicolumn{1}{c!{\vrule width 0.1pt}}{\emph{\liuhaothree{RMSE}}\liuhaothree{: 6.950}} &\multicolumn{1}{c}{\emph{\liuhaothree{RMSE}}\liuhaothree{: \textbf{2.050}}} \\ \noalign{\hrule height 0.8pt}
    Re-training from Scratch & \multicolumn{2}{c|}{\emph{\liuhaothree{ROC-AUC}}\liuhaothree{: 0.6709}} & \multicolumn{2}{c}{\emph{\liuhaothree{RMSE}}\liuhaothree{: 1.300}} \\ 
    
    \noalign{\hrule height 0.8pt}
  \end{tabular}}
  \end{center}
  \label{tab:homograph}
  \vspace{-5.6mm}
  \end{table}

  \begin{table}[b]
    \vspace{-4mm}
    \caption{Results of 3D object recognition tasks with DGCNN \cite{wang2019dynamic}.}
    \vspace{-6.5mm}
    \begin{center}
    \scriptsize
    \setlength\tabcolsep{1.4 pt}
    {\renewcommand{\arraystretch}{1}
    \begin{tabular}{l|c|c|c|>{\centering}p{1.05cm}|>{\centering\arraybackslash}p{1.05cm}}
      \noalign{\hrule height 0.8pt}
      \multirow{2}{*}{\textbf{Types}} & \multirow{2}{*}{\textbf{Tasks}} & \multirow{2}{*}{\textbf{\# Classes}} & \textbf{\emph{Pre-trained}} & \multicolumn{2}{c}{\textbf{\emph{Reusing Performance}}}  \\ 
      \cline{5-6}
      & & & \textbf{\emph{Performance}} & \multicolumn{1}{c|}{\textbf{Vanilla}} & \textbf{Ours}\\ \hline
      Pre-trained Acc & \texttt{ModelNet40} & 40 & 0.9327 &N/A & N/A \\ \noalign{\hrule height 0.2pt}
      Downstream Acc & \texttt{ShapeNet} & 16 & N/A & 0.1545 & \textbf{0.6090} \\
  
      \noalign{\hrule height 0.8pt}
    \end{tabular}}
    \end{center}
    \label{tab:pointcloud}
    \vspace{-3.5mm}
    \end{table}

\begin{figure}[t]
    \centering
    \includegraphics[width=0.475\textwidth]{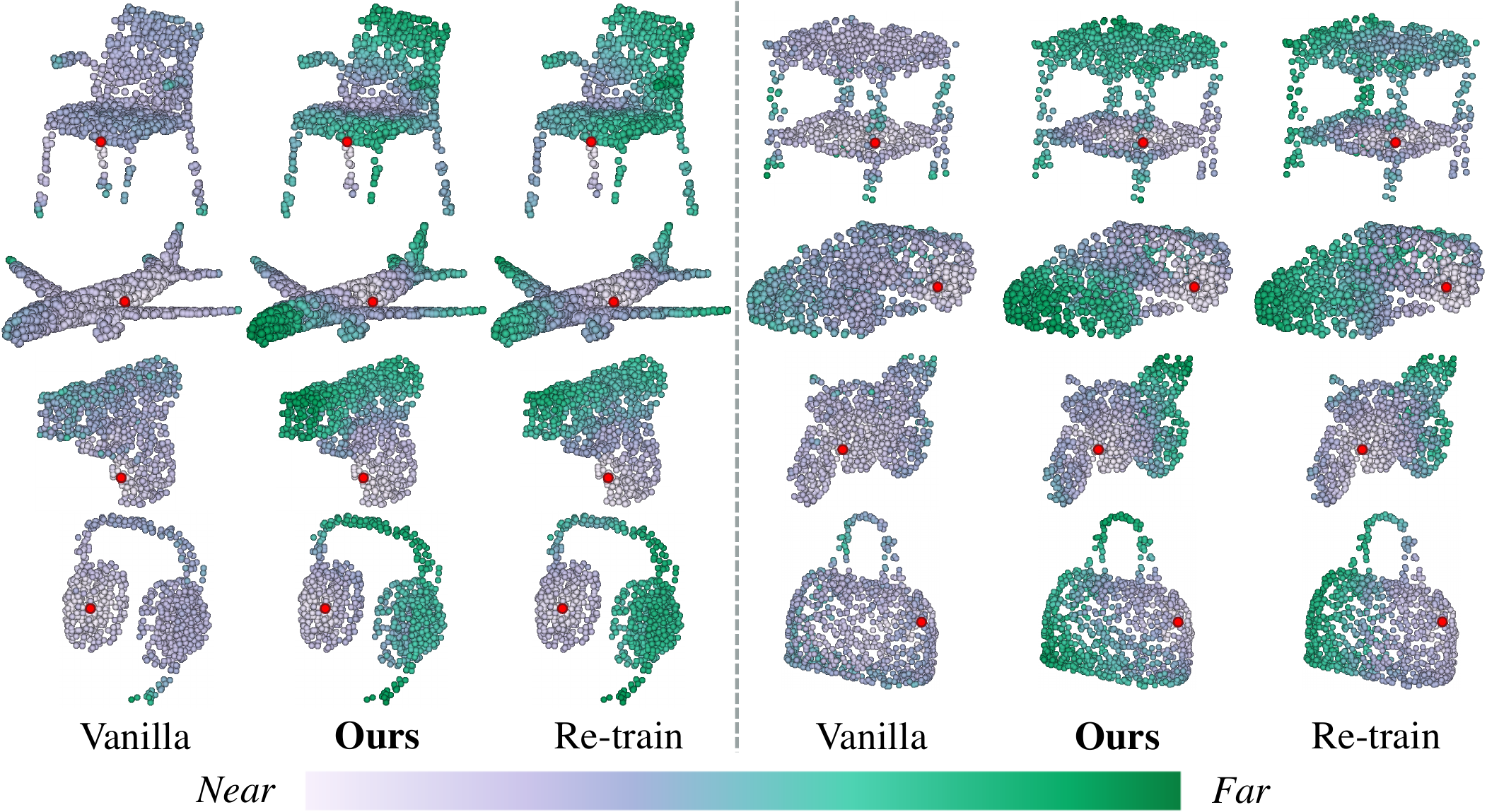}
    \vspace{-6mm}
    \caption{Visualization results of feature space structures, depicted as the distance between the red point and the rest of the others.}

    \label{fig:pointcloud}
    \vspace{-2.6mm}
  \end{figure}

\noindent\textbf{3D Object Recognition.}
Tab.~\ref{tab:pointcloud} shows the results of reusing a pre-trained DGCNN tailored for {\small\texttt{ModelNet40}} \cite{wu20153d}, to tackle distinct downstream classes in {\small\texttt{ShapeNet}} \cite{yi2016scalable}.
Remarkably, the proposed \emph{MetaGP} makes it possible for such cross-domain model reusing with large-scale 3D datasets.
We also illustrate in Fig.~\ref{fig:pointcloud} the structure of the feature space at the intermediate layer, showing that ours leads to semantically similar structures to those of re-training from scratch.

\noindent\textbf{Distributed Action Recognition.}
We construct temporally growing graphs from \texttt{WARD} \cite{yang2009distributed,wang2018kernel} and convert the problem of distributed action recognition into that of subgraph classification, as is also done in \cite{wang2022lifelong}. The results are shown in Tab.~\ref{tab:actionrecognition}, demonstrating the effectiveness of our method.

\begin{table}[H]
  \vspace{-2mm}
  \caption{Results 
  of distributed action recognition with incremental time-series data streams and categories as downstream tasks.
  }
  \vspace{-6.8mm}
  \begin{center}
  \scriptsize
  
  {\setlength\tabcolsep{0.8 pt}
    \renewcommand{\arraystretch}{1.1}
    \begin{tabular}{l|ccccccccc}

    \noalign{\hrule height 0.8pt}
  
  \multirow{2}{*}{\textbf{Tasks}} &  \multicolumn{9}{c}{\textbf{Pre-trained Action Categories}} \\ 
    &  \texttt{Up} & \texttt{ReLi} & \texttt{WaLe} & \texttt{TuLe} & \texttt{Down} & \texttt{Jog} & \texttt{Push}& \texttt{ReSt} & \textbf{Acc}\\  \hline
    \textbf{Pre-trained Acc} & 0.9721 & 0.8563 & 0.9704 & 0.9731 & 0.9265 & 0.9875 & 0.9522 & 0.9229 & 0.9366\\

  \hline
  \end{tabular}}
  {\setlength\tabcolsep{5 pt}
    \renewcommand{\arraystretch}{1}
  \begin{tabular}{p{1.8cm}|ccccccc}

    \hline
  
  \multirow{2}{*}{\textbf{Tasks}} & \multicolumn{5}{c}{\textbf{Downstream Action Categories}} \\ 
    & \texttt{ReSi} & \texttt{WaFo} & \texttt{TuRi} & \texttt{WaRi} & \texttt{Jump} & \textbf{Acc}\\ \hline
    \textbf{Downstream Acc} & 0.7337 & 0.9574 & 0.6419 & 0.6893 & 0.8081 & \textbf{0.7871} \\ 
    \noalign{\hrule height 0.8pt}
  \end{tabular}}
  \end{center}
  \label{tab:actionrecognition}
  \vspace{-5mm}
  \end{table}

\section{Conclusions}

In this paper, we introduce a novel \textsc{Gare} task for resource-efficient and generalized model reusing, tailored for GNNs.
Our objective is to reuse a pre-trained GNN for diverse cross-level/domain downstream tasks, being rid of re-training or fine-tuning.
To this end, we identified two key challenges on the data and model sides, respectively, and propose a suit of three \emph{data reprogramming} (\textsc{Dare}) and one \emph{model reprogramming} (\textsc{Mere}) approaches to resolve the dilemma.
Experiments
on fourteen benchmarks across various domains, including 
node and graph classification, graph property regression, 3D object recognition, and distributed action recognition, demonstrate that the proposed methods lead to encouraging downstream performance, and meanwhile enjoy a low computational cost.
In our future work, we will strive to generalize \textsc{Gare} to other domains.

\vspace{-1mm}
\section*{Acknowledgements}\vspace{-0.5mm}
This research is supported by Australian Research Council Projects in part by FL170100117 and IH180100002, and the National Research Foundation Singapore under its AI Singapore Programme (Award Number: AISG2-RP-2021-023).


{\small
\bibliographystyle{ieee_fullname}
\bibliography{MYRE}
}

\end{document}